\documentclass[11pt]{article} 
\usepackage{fullpage,eso-pic,ifthen}
\usepackage{amsmath,amssymb,amsthm,graphicx}
\usepackage{cite}

\newcommand\sE{\ensuremath{\mathcal{E}}}

\newcommand\sO{\ensuremath{\mathcal{O}}}
\newcommand\sP{\ensuremath{\mathcal{P}}}

\newcommand\sS{\ensuremath{\mathcal{S}}}

\newcommand\sV{\ensuremath{\mathcal{V}}}

\newcommand\bx{\ensuremath{\mathbf{x}}}

\newcommand\bP{\ensuremath{\mathbf{P}}}

\newcommand\BP{\ensuremath{\mathbb{P}}}

\newcommand\FigTop[4]{\begin{figure}[t] \begin{center} \includegraphics[scale=#2]{#1} \end{center} \caption{\label{fig:#3} #4} \end{figure}}

\DeclareMathOperator*{\diag}{diag} 
\newcommand\p[1]{\ensuremath{\left( #1 \right)}} 
\newcommand\pa[1]{\ensuremath{\left\langle #1 \right\rangle}}

\newcommand\half{\ensuremath{\frac{1}{2}}}
\newcommand\R{\ensuremath{\mathbb{R}}} 
 
\newcommand\inner[2]{\ensuremath{\left< #1, #2 \right>}}

\newcommand\eqdef{\ensuremath{\stackrel{\rm def}{=}}} 
\newcommand{\1}{\mathbb{I}} 
\newcommand{\bone}{\mathbf{1}} 
 
\newcommand\refeqn[1]{(\ref{eqn:#1})}

\newcommand\refsec[1]{Section~\ref{sec:#1}}

\newcommand\reffig[1]{Figure~\ref{fig:#1}}

\newcommand\refapp[1]{Appendix~\ref{sec:#1}}
\newcommand\refthm[1]{Theorem~\ref{thm:#1}}

\newcommand\reflem[1]{Lemma~\ref{lem:#1}}

\newcommand\Section[2]{\section{#2}\label{sec:#1}}
\newcommand\Subsection[2]{\subsection{#2}\label{sec:#1}}

\ifthenelse{\isundefined{\definition}}{\newtheorem{definition}{Definition}}{}
\ifthenelse{\isundefined{\assumption}}{\newtheorem{assumption}{Assumption}}{}
\ifthenelse{\isundefined{\proposition}}{\newtheorem{proposition}{Proposition}}{}
\ifthenelse{\isundefined{\theorem}}{\newtheorem{theorem}{Theorem}}{}
\ifthenelse{\isundefined{\lemma}}{\newtheorem{lemma}{Lemma}}{}
\ifthenelse{\isundefined{\corollary}}{\newtheorem{corollary}{Corollary}}{}

\DeclareMathOperator*{\E}{\mathbb{E}}

\newcommand\Span[2]{{[#1:#2]}}
\newcommand\OneHot[1]{\pa{#1}}

\newcommand\Aleft{A_\swarrow}
\newcommand\Aright{A_\searrow}
\newcommand\thetatrue{{\theta_0}}
\newcommand\thetasamp{{\tilde\theta}}

\newcommand\otimescol{\otimes^{\text{\sc c}}}

\renewcommand\E{\mathbb{E}}

\DeclareMathOperator*{\range}{range}
\DeclareMathOperator*{\Decompose}{\text{\sc Decompose}}
\DeclareMathOperator*{\CheckIdentifiability}{\text{\sc CheckIdentifiability}}
\DeclareMathOperator*{\Unmix}{\text{\sc Unmix}}

\DeclareMathOperator*{\Tree}{Topology}
\DeclareMathOperator*{\Trees}{Topologies}
\DeclareMathOperator*{\Root}{Root}
\DeclareMathOperator*{\dir}{dir}
\DeclareMathOperator*{\rank}{rank}

\newcommand\All{{\operatorname{all}}}
\newcommand\Pairs{{12}}
\newcommand\AllPairs{{**}}
\newcommand\Triples{{123}}
\newcommand\AllTriples{{***}}
\newcommand\ThinTriples[1]{{123#1}}
\newcommand\AllThinTriples[1]{{***#1}}

\newcommand\Lmax{L_{\operatorname{max}}}

\newcommand\eqclass[1]{\ensuremath{\sS_{#1}}}

\newcommand{\tcell}[2][c]{\begin{tabular}[#1]{@{}c@{}}#2\end{tabular}}

\newcommand\ind{\textcolor{white}{$-$}}

\DeclareMathOperator*{\StartNode}{\text{\sc Start}}
\DeclareMathOperator*{\EndNode}{\text{\sc End}}

\DeclareMathOperator*{\Combine}{\text{\sc Combine}}
\newcommand\nil{\text{\o}}

\title{Identifiability and Unmixing of Latent Parse Trees}

\author{Daniel Hsu \\ Microsoft Research \and Sham M. Kakade \\ Microsoft
Research \and Percy Liang \\ Google Research}

\begin{document}

\maketitle

\begin{abstract}
This paper explores unsupervised learning of parsing models along two directions.
First, which models are identifiable from infinite data?
We use a general technique for numerically checking identifiability based
on the rank of a Jacobian matrix, and apply it to
several standard constituency and dependency parsing models.
Second, for identifiable models, how do we estimate the parameters efficiently?
EM suffers from local optima, while
recent work using spectral methods \cite{anandkumar12moments} cannot be directly applied
since the topology of the parse tree varies across sentences.  We develop a
strategy, unmixing, which deals with this additional complexity for restricted
classes of parsing models.
\end{abstract}

\Section{introduction}{Introduction}

Generative parsing models, which define joint distributions over sentences and their parse
trees, are one of the core techniques in computational linguistics.  We
are interested in the unsupervised learning of these models
\cite{pereira92bracket,carroll92dependency,paskin02bigrams,klein02conditional,klein04induction},
where the goal is to estimate the model parameters given only examples of sentences.
Unsupervised learning can fail for a number of reasons \cite{liang08errors}:
model misspecification, 
non-identifiability, 
estimation error, 
and computation error. 
In this paper, we delve into two of these issues: identifiability and computation.
In doing so, we confront a central challenge of parsing models---that
the topology of the parse tree is unobserved
and varies across sentences.  This is in contrast to standard phylogenetic models
\cite{chang96ident}
and other latent tree models for
which there is a single fixed global tree across all examples
\cite{anandkumar11tree}.

A model is identifiable if there is enough information in the data to
pinpoint the parameters (up to some trivial equivalence class);
establishing the identifiability of a model is often a highly non-trivial
task.
A classic result of Kruskal \cite{kruskal77three} has been employed to prove
the identifiability of a wide class of latent variable models, including
hidden Markov models and certain restricted mixtures of latent tree models
\cite{allman09identifiability,allman11identifiability,rhodes12trees}.
However, these techniques cannot be directly applied
to parsing models since the tree topology varies
over an exponential set of possible topologies.
Instead, we turn to techniques from algebraic geometry
\cite{rothenberg71parametric,goodman74latent,bamber85testable,geiger01stratified}; we show
that a simple numerical procedure can be used to check identifiability
for a wide class of models in NLP.
Using this tool, we discover that probabilistic context-free grammars
(PCFGs) are non-identifiable, but that simpler PCFG variants and dependency
models are identifiable.

The most common way to estimate unsupervised parsing models is by using local
techniques such as EM \cite{lari90scfg} or MCMC sampling \cite{johnson07mcmc},
but these methods can suffer from local optima and slow mixing.
Meanwhile, recent work
\cite{mossel06hmm,hsu09spectral,siddiqi10hmm,parikh11tree,anandkumar12moments}
has shown that spectral methods can be used to estimate mixture models and HMMs
with provable guarantees.
These techniques express low-order moments of the observable distribution as a
product of matrix parameters and use eigenvalue decomposition to recover
these matrices.  However, these methods are not directly applicable to parsing models because
the tree topology again varies non-trivially.  To address this, we propose a new technique, {\em unmixing}.
The main idea is to express moments of the observable distribution as a mixture
over the possible topologies.  For restricted parsing models, the moments for a
fixed tree structure can be ``unmixed'', thereby reducing the problem to one with
a fixed topology, which can be tackled using standard techniques \cite{anandkumar12moments}.
Importantly, our unmixing technique does not require the training sentences
be annotated with the tree topologies {\em a priori}, in contrast
to recent extensions of \cite{hsu09spectral} to learning PCFGs \cite{cohen12pcfg} and
dependency trees \cite{luque12dependency,dhillon12dependency}, which work on a fixed topology.

\Section{notation}{Notation}

For a positive integer $n$, define $[n] \eqdef \{ 1, \dots, n \}$
and 
$\OneHot{n} = \{ e_1, \dots, e_n \}$,
where $e_i$ is the vector which is 1 in component $i$ and 0 elsewhere.
For integers $a, b \in [n]$,
let $a \otimes_n b = (a-1) n + b \in [n^2]$ be the integer encoding of the pair $(a,b)$.
For a pair of matrices, $A, B \in \R^{m \times n}$,
define the columnwise tensor product
$A \otimescol B \in \R^{m^2 \times n}$ to be such that $(A \otimescol B)_{(i_1
\otimes_m i_2) j} = A_{i_1 j} B_{i_2 j}$.
For a matrix $A \in \R^{m \times n}$, let $A^\dagger$ denote the Moore-Penrose
pseudoinverse.

\Section{models}{Parsing models}

A sentence is a sequence of $L$ words, $\bx
= (x_1, \dots, x_L)$, where each word $x_i \in \OneHot{d}$ is one of $d$
possible word types.
A (generative) {\em parsing model} defines a joint distribution $\BP_\theta(\bx, z)$ over
a sentence $\bx$ and its parse tree $z$ (to be made precise later),
where $\theta$ are the model parameters (a collection of multinomials).
Each parse tree $z$ has a {\em topology} $\Tree(z) \in \Trees$, which is both
unobserved and varying across sentences.
The learning problem is to recover $\theta$ given only samples of $\bx$.

Two important classes of models of natural language syntax are constituency
models, which represent a hierarchical grouping and labeling of the phrases of
a sentence (e.g., \reffig{exampleTrees}(a)), and dependency models, which represent
pairwise relationships between the words of a sentence (e.g., \reffig{exampleTrees}(b)).

\FigTop{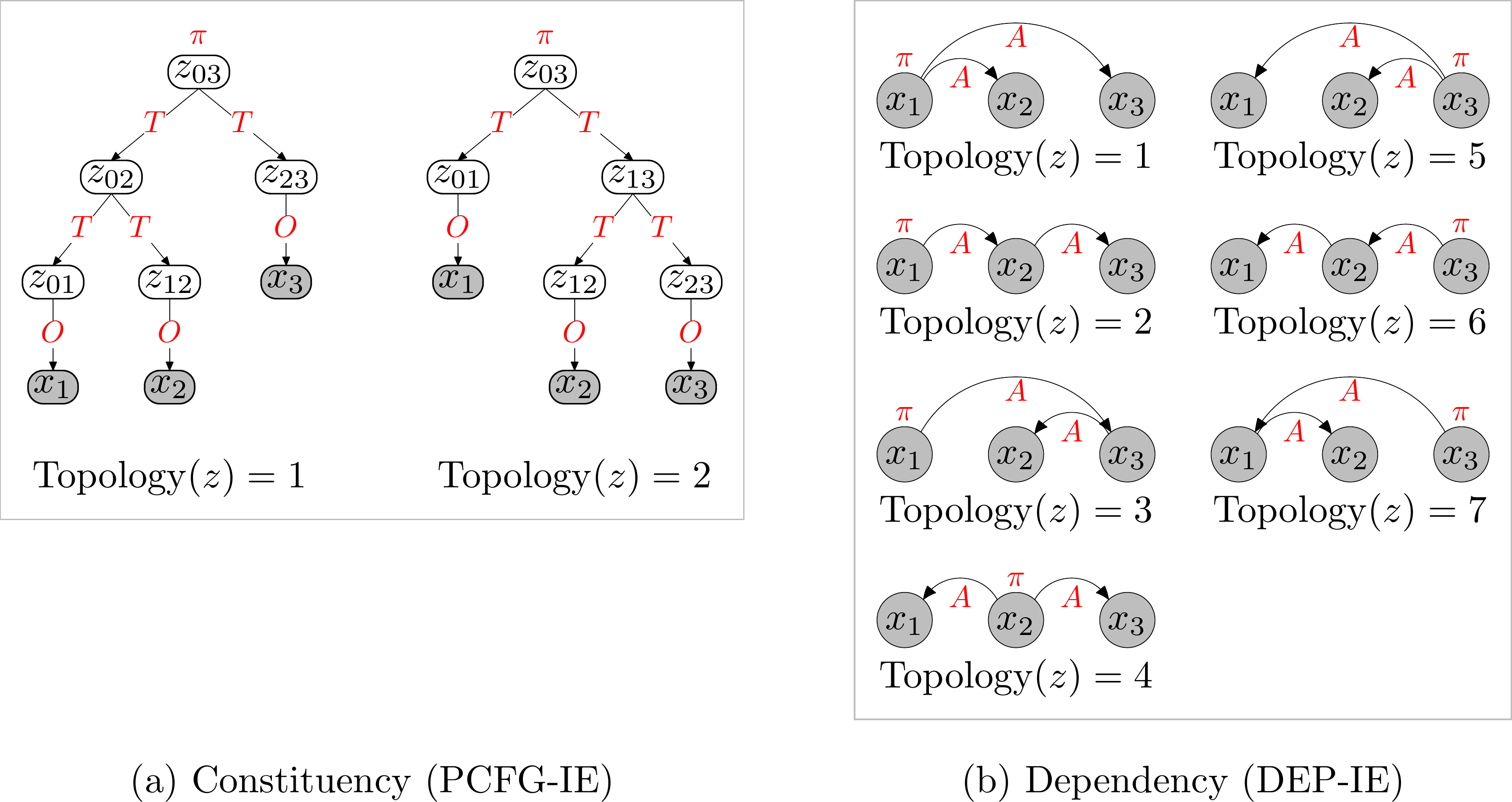}{0.40}{exampleTrees}{
The two constituency trees and seven dependency trees over $L=3$ words,
$x_1, x_2, x_3$.
(a) A constituency tree consists of a hierarchical grouping of the words
with a latent state $z_v$ for each node $v$.
(b) A dependency tree consists of a collection of directed edges between the
words.
In both cases, we have labeled each edge from $i$ to $j$ with the parameters
used to generate the state of node $j$ given $i$.
}

\Subsection{constituency}{Constituency models}

A {\em constituency tree} $z = (V, s)$ consists of a set of nodes $V$ and
a collection of hidden states $s = \{ s_v \}_{v \in V}$.
Each state $s_v \in \OneHot{k}$ represents one of $k$ possible syntactic
categories. 
Each node $v$ has the form $\Span{i}{j}$ for $0 \le i < j \le L$ corresponding
to the phrase between positions $i$ and $j$ of the sentence.
These nodes form a binary tree as follows: the root node is $\Span{0}{L} \in
V$, and for each node $\Span{i}{j} \in V$ with $j-i > 1$, there exists a unique $m$
with $i < m < j$ defining the two children nodes $\Span{i}{m} \in V$ and
$\Span{m}{j} \in V$.
Let $\Tree(z)$ be an integer encoding of $V$.

\paragraph{PCFG.}

Perhaps the most well-known constituency parsing model is the probabilistic
context-free grammar (PCFG).
The parameters of a PCFG are $\theta = (\pi, B, O)$,
where $\pi \in \R^k$ specifies the initial state distribution,
$B \in \R^{k^2 \times k}$ specifies the binary production distributions,
and $O \in \R^{d \times k}$ specifies the emission distributions.

A PCFG corresponds to the following generative process
(see \reffig{exampleTrees}(a) for an example):
choose a topology $\Tree(z)$ uniformly at random;\footnote{
Usually a PCFG induces a topology via a state-dependent probability of choosing
a binary production versus an emission.
Our model is a restriction which corresponds to a state-independent probability.}
generate the state of the root node using $\pi$;
recursively generate pairs of children states given their parents using $B$;
and finally generate words $x_i$ given their parents using $O$.
This generative process defines a joint probability over a sentence $\bx$ and a
parse tree $z$:
\begin{align}
\BP_\theta(\bx, z) = |\Trees|^{-1} \pi^\top s_{\Span{0}{L}}
\prod_{\Span{i}{m},\Span{m}{j} \in V} (s_{\Span{i}{m}} \otimes_k s_{\Span{m}{j}})^\top B s_{\Span{i}{j}}
\prod_{i=1}^L x_i^\top O s_{\Span{i-1}{i}},
\end{align}

We will also consider two variants of the PCFG with additional restrictions:

\paragraph{PCFG-I.}
The left and right children states are generated independently---that is, we
have the following factorization: $B = T_1 \otimescol T_2$ for some $T_1,T_2 \in
\R^{k \times k}$.

\paragraph{PCFG-IE.}
The left and the right productions are independent and equal: $B = T \otimescol T$.

\Subsection{dependency}{Dependency tree models}

In contrast to constituency trees, which posit internal nodes with latent
states, dependency trees connect the words directly.  A {\em dependency
tree} $z$ is a set of directed edges $(i, j)$, where $i, j \in [L]$ are
distinct positions in the sentence.  Let $\Root(z)$ denote the position of
the root node of $z$.  We consider only {\em projective} dependency trees
\cite{eisner96dependency}: $z$ is projective if for every path
from $i$ to $j$ to $k$ in $z$,
we have that
$j$ and $k$ are on the same side of $i$ (that is, $j-i$ and $k-i$ have the same sign).
Let $\Tree(z)$ be an integer encoding of $z$. 

\paragraph{DEP-I.}

We consider the simple dependency model of \cite{paskin02bigrams}.
The parameters of this model are $\theta = (\pi, \Aleft, \Aright)$, where $\pi \in \R^d$ is
the initial word distribution and $\Aleft, \Aright \in \R^{d \times d}$ are the
left and right argument distributions. 
The generative process is as follows:
choose a topology $\Tree(z)$ uniformly at random,
generate the root word using $\pi$,
and recursively generate argument words to the left to the right given the parent word
using $\Aleft$ and $\Aright$, respectively.
The corresponding joint probability distribution is as follows:
\begin{align}
\BP_\theta(\bx, z) = |\Trees|^{-1} \pi^\top x_{\Root(z)} \prod_{(i,j) \in z} x_j^\top A_{\dir(i,j)} x_i,
\end{align}
where $\dir(i,j) = \,\swarrow$ if $j<i$ and $\searrow$ if $j>i$.

We also consider the following two variants:

\paragraph{DEP-IE.}
The left and right argument distributions are equal: $A = \Aleft = \Aright$.

\paragraph{DEP-IES.}
$A = \Aleft = \Aright$ and $\pi$ is the stationary distribution of $A$ (that is, $\pi = A \pi$).

\Section{identifiability}{Identifiability}

Our goal is to estimate model parameters $\thetatrue \in \Theta$ given only access to
sentences $\bx \sim \BP_{\thetatrue}$.  Specifically,
suppose we have an {\em observation function} $\phi(\bx) \in \R^m$,
which is the only lens through which an algorithm can view the data.
We ask a basic question:
in the limit of infinite data, is it information-theoretically possible to
identify $\thetatrue$ from the {\em observed moments} $\mu(\thetatrue) \eqdef
{\E}_{\thetatrue}[\phi(\bx)]$?

To be more precise,
define the {\em equivalence class} of $\thetatrue$ to be the set of parameters
$\theta$ that yield the same observed moments:
\begin{align}
\label{eqn:equiv}
\eqclass{\Theta}(\thetatrue) &= \{ \theta \in \Theta : \mu(\theta) = \mu(\thetatrue) \}.
\end{align}
It is impossible for an algorithm to distinguish among the elements of
$\eqclass{\Theta}(\thetatrue)$.  Therefore, one might want to ensure that
$|\eqclass{\Theta}(\thetatrue)|
= 1$
for all $\thetatrue \in \Theta$.  However, this requirement is too strong for two reasons.
First, models often have natural symmetries---e.g., the $k$ states of any PCFG
can be permuted without changing $\mu(\theta)$, so $|\eqclass{\Theta}(\thetatrue)| \ge k!$.
Second, $|\eqclass{\Theta}(\thetatrue)| = \infty$ for some pathological $\thetatrue$'s---e.g., 
PCFGs where all states have the same emission distribution $O$
are indistinguishable regardless of the production distributions $B$.
The following definition of identifiability accommodates these two exceptional cases:

\begin{definition}[Identifiability]
\label{def:identifiability}
A model family with parameter space $\Theta$ is {\em
(globally) identifiable from $\phi$} if there exists a measure zero set $\sE$ such that
$|\eqclass{\Theta}(\thetatrue)|$ is finite for every
$\thetatrue \in \Theta\backslash\sE$.
It is \emph{locally identifiable from $\phi$} if there exists a measure
zero set $\sE$ such that, for every $\thetatrue \in \Theta\backslash\sE$,
there exists an open neighborhood $N(\thetatrue)$ around $\thetatrue$ such
that $\eqclass{\Theta}(\thetatrue) \cap N(\thetatrue) =
\{\thetatrue\}$.
\end{definition}

\paragraph{Example of non-identifiability.}
Consider the DEP-IE model with $L=2$ with the full observation function
$\phi(\bx) = x_1 \otimes x_2$.
The corresponding observed moments are $\mu(\theta) = 0.5 A \diag(\pi) + 0.5 \diag(\pi) A^\top$.
Note that $A \diag(\pi)$ is an arbitrary $d \times d$ matrix whose entries sum to 1,
which has $d^2-1$ degrees of freedom, whereas $\mu(\theta)$ is a symmetric matrix
whose entries sum to 1, which has $\binom{d+1}{2} - 1$ degrees of freedom.
Therefore, $\eqclass{\Theta}(\theta)$ has dimension $\binom{d}{2}$ and therefore
the model is non-identifiable.
\vspace{-.0cm}

\paragraph{Parameter counting.}
It is important to compute the degrees of freedom correctly---simple parameter
counting is insufficient.  For example, consider the PCFG-IE model with
$L=2$.  The observed moments with respect to $\phi(\bx) = x_1 \otimes
x_2$ is a $d \times d$ matrix, which places $d^2$ constraints on the $k^2 +
(d-1)k$ parameters.  When $d \ge 2k$, there are more constraints than
parameters, but the PCFG-IE model with $L=2$ is
actually non-identifiable (as we will see later).
The issue here is that the number of constraints
does not reveal the fact that some of these constraints are redundant.
\vspace{-.0cm}

\Subsection{moments}{Observation functions}

An observation function $\phi(\bx)$ and
its associated observed moments $\mu(\thetatrue) = \E_{\thetatrue}[\phi(\bx)]$
reveals aspects of the distribution $\BP_{\thetatrue}(\bx)$.
For example, $\phi(\bx) = x_1$ would only reveal the marginal distribution of the first word,
whereas $\phi(\bx) = x_1 \otimes \cdots \otimes x_L$ reveals the entire distribution of $\bx$.
There is a tradeoff:~Higher-order moments provide more information, but are
harder to estimate reliably given
finite data, and are also computationally more expensive.
In this paper, we consider the following intermediate moments:
\begin{align*}
\phi_{\Pairs}(\bx)
& \eqdef x_1 \otimes x_2
& \phi_{\AllPairs}(\bx)
& \eqdef \bigl( x_i \otimes x_j : i, j \in [L] \bigr) \\
\phi_{\Triples}(\bx)
& \eqdef x_1 \otimes x_2 \otimes x_3
& \phi_{\AllTriples}(\bx)
& \eqdef \bigl( x_i \otimes x_j \otimes x_k : i, j, k \in [L] \bigr) \\
\phi_{\ThinTriples{\eta}}(\bx)
& \eqdef (x_1 \otimes x_2) (\eta^\top x_3)
& \phi_{\AllThinTriples{\eta}}(\bx)
& \eqdef \bigl( (x_i \otimes x_j) (\eta^\top x_k)
: i,j,k \in [L] \bigr) \\
\phi_\All(\bx)
& \eqdef x_1 \otimes \cdots \otimes x_L
&
&
\end{align*}
Above, $\eta \in \R^d$ denotes a unit vector in $\R^d$ (e.g.,
$e_1$) which picks out a linear combination of
matrix slices from a third-order $d \times d \times d$ tensor.

\Subsection{identifiabilityChecker}{Automatically checking identifiability}

One immediate goal is to determine which models in \refsec{models} are
identifiable from which of the observed moments (\refsec{moments}).
A powerful analytic tool that has been succesfully applied in previous work
is Kruskal's theorem \cite{kruskal77three,allman09identifiability}, but
(i) it is does not immediately apply to models with random topologies, and
(ii) only gives sufficient conditions for identifiability, and cannot be
used to determine non-identifiability.
Furthermore, since it is common practice to explore many different
models for a given problem in rapid succession, we would like to check identifiability
quickly and reliably.  In this section, we develop an automatic procedure to do this.

To establish identifiability, let us examine the algebraic structure of
$\eqclass{\Theta}(\thetatrue)$ for $\thetatrue \in \Theta$, where we assume that
the parameter space $\Theta$ is an open subset of
$[0,1]^n$.\footnote{While we initially defined $\theta$ to be a tuple of conditional
probability matrices, we will now use its non-redundant vectorized form
$\theta \in \R^n$.}
Recall that $\eqclass{\Theta}(\thetatrue)$ is defined by the moment constraints
$\mu(\theta) = \mu(\thetatrue)$.
We can write these constraints as $h_\thetatrue(\theta) = 0$, where
\begin{align*}
h_\thetatrue(\theta) \eqdef \mu(\theta) - \mu(\thetatrue)
\end{align*}
is a vector of $m$ polynomials in $\theta$.

Let us now compute the number of degrees of
freedom of $h_\thetatrue$ around $\thetatrue$.  The key quantity is
$J(\theta) \in \R^{m \times n}$, the Jacobian of $h_{\thetatrue}$ at
$\theta$ (note that the Jacobian of $h_\thetatrue$ does not depend on
$\thetatrue$; it is precisely the Jacobian of $\mu$).
This Jacobian criterion is well-established in algebraic geometry, and has
been adopted in the statistical literature for testing model
identifiability and other related properties
\cite{rothenberg71parametric,goodman74latent,bamber85testable,geiger01stratified}.

Intuitively, each row of $J(\thetatrue)$ corresponds to a direction of a
constraint violation, and thus the row space of $J(\thetatrue)$ corresponds to
all directions that would take us outside the equivalence class $\eqclass{\Theta}(\thetatrue)$.  If
$J(\thetatrue)$ has less than rank $n$, then there is a direction orthogonal to
all the rows along which we can move and still satisfy all the constraints---in
other words, $|\eqclass{\Theta}(\thetatrue)|$ is infinite, and therefore the model is
non-identifiable.  This intuition leads to the following algorithm:
\vspace{-.0cm}

\begin{center} \scalebox{0.95}{\framebox{ \begin{minipage}{5in} 
$\CheckIdentifiability$: \\
\ind 1. Choose a point $\thetasamp \in \Theta$ uniformly at random. \\
\ind 2. Compute the Jacobian matrix $J(\thetasamp)$. \\
\ind 3. Return ``yes'' if the rank of $J(\thetasamp) = n$ and
``no'' otherwise.
\end{minipage} }} \end{center} 
\vspace{-.0cm}

The following theorem asserts the correctness of
$\CheckIdentifiability$.
It is largely based on techniques in \cite{bamber85testable}, although we
have not seen it explicitly stated in this form.
\begin{theorem}[Correctness of $\CheckIdentifiability$] \label{thm:checker}
Assume the parameter space $\Theta$ is a non-empty open connected subset of
$[0,1]^n$; and the observed moments $\mu
\colon \R^n \to \R^m$, with respect to observation function $\phi$, is a polynomial map.
Then with probability 1,
$\CheckIdentifiability$ returns ``yes'' iff the model family is
locally identifiable from $\phi$.
Moreover, if it returns ``yes'', then there exists $\sE \subset
\Theta$ of measure zero such that the model family with parameter space
$\Theta \setminus \sE$ is identifiable from $\phi$.
\end{theorem}
The proof of \refthm{checker} is given in \refapp{proofs}.

\Subsection{implementation}{Implementation of $\CheckIdentifiability$}

\paragraph{Computing the Jacobian.}
The rows of $J$ correspond to $\partial \E_\theta[\phi_j(\bx)] / \partial \theta$ and can be
computed efficiently by adapting dynamic programs used in the E-step of an EM
algorithm for parsing models.  There are two main differences: (i) we must sum over possible
values of $\bx$ in addition to $z$, and (ii) we are not computing moments, but rather gradients
thereof.  Specifically, we adapt the CKY algorithm for constituency models and
the algorithm of \cite{eisner96dependency} for dependency models.
See \refapp{dynamicProgramJacobian} for more details.

\paragraph{Numerical issues.}
Because we implemented $\CheckIdentifiability$ on a finite precision machine,
the results are subject to
numerical precision errors.
However, we verified that our numerical results are consistent with various analytically-derived identifiability results (e.g.,
from \cite{allman09identifiability}).

\Subsection{results}{Identifiability of constituency and dependency
tree models}

We checked the identifiability status of various constituency and
dependency tree models using our implementation of $\CheckIdentifiability$.
We focus on the regime where $d \geq k$ for PCFGs; additional results for
$d < k$ are given in \refapp{more}.

\begin{figure}
\begin{center}
\begin{tabular}{c|c|c|c|c|c|c|}
\cline{2-7}
Model $\backslash$ Observation function & $\phi_\Pairs$ & $\phi_\AllPairs$ & $\phi_\ThinTriples{e_1}$ &
$\phi_\Triples$ & $\phi_\AllThinTriples{e_1}$ &
$\phi_\AllTriples$
\\
\hline
\multicolumn{1}{|c|}{PCFG} &
\multicolumn{6}{|c|}{No, even from $\phi_\All$ for $L \in \{3,4,5\}$} \\
\hline
\multicolumn{1}{|c|}{PCFG-I / PCFG-IE} &
No &
Yes iff $L \geq 4$ &
\multicolumn{4}{|c|}{Yes iff $L \geq 3$} \\
\hline
\multicolumn{1}{|c|}{DEP-I} &
No &
\multicolumn{5}{|c|}{Yes iff $L \geq 3$} \\
\hline
\multicolumn{1}{|c|}{DEP-IE / DEP-IES} &
\multicolumn{6}{|c|}{Yes iff $L \geq 3$} \\
\hline
\end{tabular}
\end{center}
\caption{Local identifiability of parsing models.
These findings are given by $\CheckIdentifiability$ have the semantics from
\refthm{checker}.
These were checked for $d \in \{2,3,\dotsc,8\}$, $k \in \{2,\dotsc,d\}$
(applies only for PCFG models), $L \in \{2,3,\dotsc,9\}$.
}
\label{fig:results}
\end{figure}

The results are reported in \reffig{results}.
First, we found that the PCFG is not identifiable from
$\phi_\All$ (and therefore not identifiable from any $\phi$)
for $L \in \{3,4,5\}$; we believe that the same holds for
all $L$.
This negative result motivates exploring restricted subclasses of
PCFGs, such as PCFG-I and PCFG-IE, which factorize the binary productions.\footnote{Note that these subclasses occupy measure zero subsets
of the PCFG parameter space, which is expected given the non-identifiability of the general
PCFG.}
For these classes, we found that the sentence length $L$ and choice of
observation function can influence identifiability: 
Both models are identifiable for large enough $L$ (e.g.,
$L \geq 3$) and with a sufficiently rich observation function (e.g.,
$\phi_\ThinTriples{\eta}$).

The dependency models, DEP-I and DEP-IE, were
all found to be identifiable for $L \geq 3$ from second-order moments $\phi_\AllPairs$.
The conditions for identifiability are less stringent than their constituency
counterparts (PCFG-I and PCFG-IE), which is natural since dependency models are simpler without the latent states.
Note that in all identifiable models, second-order moments suffice
to determine the distribution---this is good news because low-order moments are
easier to estimate.

\Section{estimators}{Unmixing algorithms}

Having established which parsing models are identifiable, we now turn to
parameter estimation for these models.  We will consider algorithms based on
moment matching---those that try to find a $\theta$ satisfying
$\mu(\theta) = u$ for some $u$.  Typically, $u$ is an empirical estimate of
$\mu(\thetatrue) = \E_{\thetatrue}[\phi(\bx)]$ based on samples $\bx \sim
\BP_{\thetatrue}$.\footnote{We will develop our algorithms assuming true moments ($u = \mu(\thetatrue)$).
The empirical moments converge to the true moments at $O_p(n^{-\half})$,
and matrix perturbation arguments (e.g., \cite{anandkumar12moments}) can be used
derive sample complexity arguments for the parameter error.}

In general, solving $\mu(\theta) = u$ corresponds to finding solutions
to systems of
multivariate polynomials, which is NP-hard~\cite{sahni74qp}.  However, $\mu(\theta)$ often has
additional structure which we can exploit.  For instance, for an HMM, the
sliced third-order moments $\mu_{123\eta}(\theta)$ can be written as a
product of parameter matrices in $\theta$, and each matrix can be recovered
by decomposing the product \cite{anandkumar12moments}.

For parsing models, the challenge is that the topology is random, so the
moments is not a single product, but a mixture over products.  To deal with
this complication, we propose a new technique, which we call {\em unmixing}:~We
``unmix'' the products from the mixtures, essentially reducing the problem to one
with a fixed topology.

We will first present the general idea of unmixing (\refsec{unmixGeneral})
and then apply it to the PCFG-IE model (\refsec{unmixPcfg})
and the DEP-IES model (\refsec{unmixDep}).

\Subsection{unmixGeneral}{General case}

We assume the observation function $\phi(\bx)$ consists of a collection
of observation matrices $\{ \phi_o(\bx) \}_{o \in \sO}$
(e.g., for $o = (i,j)$, $\phi_o(\bx) = x_i \otimes x_j$).
Given an observation matrix $\phi_o(\bx)$ and a topology $t \in \Trees$,
consider the mapping that computes the observed moment conditioned on that topology:
$\Psi_{o,t}(\theta) = {\E}_\theta[\phi_o(\bx) \mid \Tree = t]$.
As we range $o$ over $\sO$ and $t$ over $\Trees$, we will enounter
a finite number of such mappings.  We call these mappings {\em compound parameters},
denoted $\{ \Psi_p \}_{p \in \sP}$.

Now write the observed moments as a weighted sum:
\begin{align}
\label{eqn:mixing}
\mu_o(\theta)
&= \sum_{p \in \sP} \underbrace{\BP(\Psi_{o,\Tree} = \Psi_p)}_{\eqdef M_{op}} \Psi_p \quad \text{for all $o \in \sO$},
\end{align}
where we have defined $M_{op}$ to be the probability mass over tree topologies
that yield compound parameter $\Psi_p$.
We let $\{ M_{op} \}_{o \in \sO, p \in \sP}$ be the {\em mixing matrix}.
Note that \refeqn{mixing} defines a system of equations $\mu = M \Psi$,
where the variables are the compound parameters and the constraints are the
observed moments.  In a sense, we have replaced the original system of
polynomial equations (in $\theta$) with a system of linear equations (in $\Psi$).

The key to the utility of this technique is that the number of compound
parameters can be polynomial in $L$ even when the number of possible
topologies is exponential in $L$.
Previous analytic techniques \cite{rhodes12trees} based on
Kruskal's theorem \cite{kruskal77three} cannot be applied here because the possible topologies
are too many and too varied.

Note that the mixing equation $\mu = M \Psi$ holds for each sentence length $L$, but many
compound parameters $p$ appear in the equations of multiple $L$.  Therefore, we can
combine the equations across all observed sentence lengths, yielding a more
constrained system than if we considered the equations of each $L$ separately.

The following proposition shows how we can recover $\theta$ by unmixing the
observed moments $\mu$:

\begin{proposition}[Unmixing]
Suppose that there exists an efficient base algorithm to recover $\theta$ from
some subset of compound parameters $\{ \Psi_p(\theta) : p \in \sP_0 \}$,
and that $e_{p}^\top$ is in the row space of
$M$ for each $p \in \sP_0$.  Then we can recover $\theta$ as follows:
\end{proposition}
\begin{center} \scalebox{0.95}{\framebox{ \begin{minipage}{5in} 
$\Unmix(\mu)$: \\
\ind 1. Compute the mixing matrix $M$ \refeqn{mixing}. \\
\ind 2. Retrieve the compound parameters $\Psi_p(\theta) = (M^\dagger \mu)_p$ for each $p \in \sP_0$. \\
\ind 3. Call the base algorithm on $\{ \Psi_p(\theta) : p \in \sP_0 \}$ to obtain $\theta$.
\end{minipage} }} \end{center} 

For all our parsing models, $M$ can be computed efficiently using dynamic
programming (\refapp{dynamicProgramMixing}).  Note that $M$ is data-independent,
so this computation can be done once in advance.

\Subsection{unmixPcfg}{Application to the PCFG-IE model}

As a concrete example, consider the PCFG-IE model over $L=3$ words.
Write $A = OT$.
For any $\eta \in \R^d$, we can express the observed moments as a sum over
the two possible topologies in \reffig{exampleTrees}(a):
\begin{align}
\mu_{123\eta} &\eqdef \E[x_1 \otimes x_2 (\eta^\top x_3)] = 0.5 \Psi_{1;\eta} + 0.5 \Psi_{2;\eta}, \quad\quad \Psi_{1;\eta} \eqdef A \diag(T \diag(\pi) A^\top\eta) A^\top, \nonumber \\
\mu_{132\eta} &\eqdef \E[x_1 \otimes x_3 (\eta^\top x_2)] = 0.5 \Psi_{3;\eta} + 0.5 \Psi_{2;\eta}, \quad\quad \Psi_{2;\eta} \eqdef A \diag(\pi) T^\top \diag(A^\top\eta) A^\top, \nonumber \\
\mu_{231\eta} &\eqdef \E[x_2 \otimes x_3 (\eta^\top x_1)] = 0.5 \Psi_{3;\eta} + 0.5 \Psi_{1;\eta}, \quad\quad \Psi_{3;\eta} \eqdef A \diag(A^\top\eta) T \diag(\pi) A^\top, \nonumber
\end{align}
or compactly in matrix form:
\begin{align}
\underbrace{\p{\begin{array}{c} \mu_{123\eta} \\ \mu_{132\eta} \\ \mu_{231\eta} \end{array}}}_{\text{observed moments $\mu_{\eta}$}}
= \underbrace{\p{\begin{array}{ccc} 0.5I & 0.5I & 0 \\ 0 & 0.5I & 0.5I \\ 0.5I & 0 & 0.5I \end{array}}}_{\text{mixing matrix $M$}}
\underbrace{\p{\begin{array}{c} \Psi_{1;\eta} \\ \Psi_{2;\eta} \\ \Psi_{3;\eta} \end{array}}}_{\text{compound parameters $\Psi_\eta$}}.
\nonumber
\end{align}
Let us observe $\mu_\eta$ at two different values of $\eta$,
say at $\eta=\bone$ and $\eta=\tau$ for some random $\tau$.
Since the mixing matrix $M$ is invertible, we can obtain
the compound parameters
$\Psi_{2;\bone} = (M^{-1} \mu_1)_2$ and $\Psi_{2;\tau} = (M^{-1} \mu_\tau)_2$.

Now we will recover $\theta$ from $\Psi_{2;\bone}$ and
$\Psi_{2;\tau}$ by first extracting $A = OT$ via an eigenvalue
decomposition, and then recovering $\pi$, $T$, and $O$ in turn (all up to the same
unknown permutation) via elementary matrix operations.

For the first step, we will use the following tool (adapted from Algorithm A of
\cite{anandkumar12moments}), which allow us to decompose two related matrix
products:
\begin{lemma}[Spectral decomposition]
\label{lem:decompose}
Let $M_1, M_2 \in \R^{d \times k}$ have full column rank and $D$ be a diagonal
matrix with distinct diagonal entries.
Suppose we observe $X = M_1 M_2^\top$ and $Y = M_1 D M_2^\top$.  Then
$\Decompose(X,Y)$ recovers $M_1$ up to a permutation and scaling of the
columns.
\end{lemma}

\begin{center} \scalebox{0.95}{\framebox{ \begin{minipage}{5in} 
$\Decompose(X,Y)$: \\
\ind 1. Find $U_1, U_2 \in \R^{d \times k}$ such that $\range(U_1) = \range(X)$
and $\range(U_2) = \range(X^\top)$. \\
\ind 2. Perform an eigenvalue decomposition of $(U_1^\top Y U_2) (U_1^\top X U_2)^{-1} = V S V^{-1}$. \\
\ind 3. Return $(U_1^\top)^\dagger V$.
\end{minipage} }} \end{center} 

First, run $\Decompose(X = \Psi_{2;\bone}^\top, Y =
\Psi_{2;\tau}^\top)$ (\reflem{decompose}),
which corresponds to $M_1 = A$ and $M_2 = A \diag(\pi) T^\top$.
This produces $A \Pi S$ for some permutation matrix $\Pi$ and diagonal scaling $S$.  Since
we know that the columns of $A$ sum to one, we can identify $A \Pi$.

To recover the initial distribution $\pi$ (up to permutation),
take $\Psi_{2;\bone} \bone = A\pi$
and left-multiply by $(A \Pi)^\dagger$ to get $\Pi^{-1} \pi$.
For $T$, put the entries of $\pi$ in a diagonal matrix: $\Pi^{-1} \diag(\pi) \Pi$.
Take $\Psi_{2;\bone}^\top = A T \diag(\pi) A^\top$
and multiply by
$(A \Pi)^\dagger$ on the left and
$((A \Pi)^\top)^\dagger (\Pi^{-1} \diag(\pi) \Pi)^{-1}$ on the right,
which yields $\Pi^{-1} T \Pi$.
(Note that $\Pi$ is orthogonal, so $\Pi^{-1} = \Pi^\top$.)
Finally, multiply $A\Pi = OT\Pi$ and $(\Pi^{-1} T \Pi)^{-1}$,
which yields $O \Pi$.

The above algorithm identifies the PCFG-IE from only length 3 sentences.  To exploit
sentences of different lengths, we
can compute a mixing matrix $M$ which includes constraints from sentences of length $1 \le L \le
\Lmax$ up to some upper bound $\Lmax$.
For example, $\Lmax = 10$ results in a $990 \times 2376$ mixing matrix.
We can retrieve the same compound parameters ($\Psi_{2;\bone}$ and $\Psi_{2;\tau}$)
from the pseudoinverse of $M$ and as proceed as before.

\Subsection{unmixDep}{Application to the DEP-IES model}

We now turn to the DEP-IES model over $L=3$ words.
Our goal is to recover the parameters $\theta = (\pi, A)$.
Let $D = \diag(\pi) = \diag(A \pi)$, where the second equality is due to
stationarity of $\pi$.
\begin{align}
\mu_1 &\eqdef \E[x_1] = \pi,
\nonumber \\
\mu_{12} &\eqdef \E[x_1 \otimes x_2] = 7^{-1} (D A^\top + D A^\top + D A^\top A^\top + A D + A D A^\top + A D + D A^\top),
\nonumber \\
\mu_{13} &\eqdef \E[x_1 \otimes x_3] = 7^{-1} (D A^\top + D A^\top A^\top + D A^\top + A D A^\top + A D + A A D + A D),
\nonumber \\
\tilde\mu_{12} &\eqdef \tilde \E[x_1 \otimes x_2] = 2^{-1} (DA^\top + A D),
\nonumber
\end{align}
where $\tilde\E[\cdot]$ is taken with respect to length 2 sentences.
Having recovered $\pi$ from $\mu_1$, it remains to recover $A$.
By selectively combining the moments above, we can compute
$AA + A = [7 (\mu_{13} - \mu_{12}) + 2 \tilde \mu_{12}] \diag(\mu_1)^{-1}$.
Assuming $A$ is generic position, it is diagonalizable:
$A = Q \Lambda Q^{-1}$ for some diagonal matrix $\Lambda =
\diag(\lambda_1,\dotsc,\lambda_d)$, possibly with complex entries.
Therefore, we can recover $\Lambda^2 + \Lambda = Q^{-1} (AA + A) Q$.
Since $\Lambda$ is diagonal, we simply have $d$ independent quadratic equations
in $\lambda_i$, which can be solved in closed form. 
After obtaining $\Lambda$, we retrieve $A = Q \Lambda Q^{-1}$.

\Section{discussion}{Discussion}

In this work, we have shed some light on the identifiability of standard
generative parsing models using our numerical identifiability checker.  Given
the ease with which this checker can be applied, we believe it should be a useful
tool for analyzing more sophisticated models \cite{klein04induction}, as well as
developing new ones which are expressive yet identifiable.

There is still a large gap between showing identifiability and developing
explicit algorithms.  We have made some progress on closing it with our
unmixing technique, which can deal with models where the tree topology varies
non-trivially.

\subsubsection*{References}
{\def\section*#1{} \bibliography{abbreviated} \bibliographystyle{unsrt}}

\appendix

\Section{proofs}{Proof of \refthm{checker}}
\newcommand\Thetamax{\ensuremath{\Theta_{\operatorname{max}}}}
\newcommand\rmax{\ensuremath{r_{\operatorname{max}}}}
\newcommand\Thetadef{\ensuremath{\Theta_{\operatorname{deficient}}}}
\newcommand\Thetaexcept{\ensuremath{\Theta_{\operatorname{except}}}}

\newtheorem*{theoremOne}{Theorem 1 (restated)}
\begin{theoremOne}
Assume $\Theta$ is a non-empty open connected subset of $[0,1]^n$ and $\mu
\colon \R^n \to \R^m$ is a polynomial map.
With probability 1, the following holds.
\begin{itemize}
\item $\CheckIdentifiability$ returns ``no'' $\Rightarrow$
for almost all $\thetatrue \in \Theta$ and any open neighborhood
$N(\thetatrue)$ around $\thetatrue$, $|\eqclass{\Theta}(\thetatrue) \cap
N(\thetatrue)|$ is infinite (not locally identifiable).

\item $\CheckIdentifiability$ returns ``yes'' $\Rightarrow$
(i) for almost all $\thetatrue \in \Theta$, there exists an open
neighborhood $N(\thetatrue)$ around $\thetatrue$ such that
$|\eqclass{\Theta}(\thetatrue) \cap N(\thetatrue)| = 1$ (locally identifiable);
and (ii) there exists a set $\sE \subset \Theta$ with measure zero such that
$|\eqclass{\Theta\setminus\sE}(\thetatrue)|$ is finite for every
$\thetatrue \in \Theta\setminus\sE$ (identifiability of $\Theta\backslash\sE$).

\end{itemize}
\end{theoremOne}

The proof of \refthm{checker} crucially relies on the following lemma from
\cite{bamber85testable} which holds even in the case that $\mu$ is merely
an analytic function (see Lemma 9 of \cite{geiger01stratified} for a
simpler proof in the case $\mu$ is a polynomial map); it states that the
Jacobian achieves its maximal rank almost everywhere in $\Theta$.
To state this precisely, first define $\rmax \eqdef \max \{
\rank(J(\theta)) : \theta \in \Theta \}$ and $\Thetamax \eqdef \{ \theta
\in \Theta : \rank(J(\theta)) = \rmax \}$.
\begin{lemma} \label{lem:max-rank}
The set $\Theta \setminus \Thetamax$ has Lebesgue measure zero.
That is, $\Thetamax$ is almost all of $\Theta$.
\end{lemma}

\begin{proof}[Proof of \refthm{checker}]
By \reflem{max-rank}, $\CheckIdentifiability$ chooses a point $\thetasamp
\in \Thetamax$ with probability 1.
We henceforth condition on this event, so $\rank(J(\thetasamp)) = \rmax$.

{\em Case 1}: $\rank(J(\thetasamp)) < n$ (\emph{i.e.}, ``no''
is returned).
In this case, we have $\rmax < n$.
We now employ an argument from the proof of Proposition 20 of
\cite{bamber85testable}.
Fix any $\thetatrue \in \Thetamax$.
Since $\Theta$ is open, Weyl's theorem implies that there is an open
neighborhood $U$ around $\thetatrue$ in $\Theta$ on which $\rank(J(\theta))
= \rmax$ for all $\theta \in U$ (\emph{i.e.}, $\rank(J(\cdot))$ is constant
on $ U$).
Therefore, by the constant rank theorem, there is an open neighborhood
$N(\thetatrue)$
around $\thetatrue$ in $\Theta$ such that $\mu^{-1}(\mu(\thetatrue)) \cap
N(\thetatrue)$ is homeomorphic with an open set in $\R^{n-\rmax}$.
Therefore $\eqclass{\Theta}(\thetatrue) \cap N(\thetatrue)$ is uncountably
infinite.

{\em Case 2}: $\rank(J(\thetasamp)) = n$ (\emph{i.e.}, ``yes'' is
returned).
In this case, we have $\rmax = n$.
Therefore for every $\thetatrue \in \Thetamax$, the Jacobian
$J(\thetatrue)$ has full column rank, and thus by the inverse function
theorem, $\mu$ is injective on a neighborhood of $\thetatrue$.
This in turn implies that for all $\thetatrue \in \Thetamax$, there exists
an open neighborhood $N(\thetatrue)$ around $\thetatrue$ such that
$\eqclass{\Theta}(\thetatrue) \cap N(\thetatrue) = \{ \thetatrue \}$.
This proves (i).

To show (ii), define $\sE \eqdef \Theta \setminus \Thetamax$, and now claim
that for every $\thetatrue \in \Thetamax$, the
equivalence class $\eqclass{\Thetamax}(\thetatrue)$ is finite.
Observe that by (i),
the set $\eqclass{\Thetamax}(\thetatrue)$ contains only geometrically isolated
solutions to the system of polynomial equations given by $\mu(\theta) =
\mu(\thetatrue)$.
Therefore the claim follows immediately from B\'ezout's Theorem, which
implies that the number of geometrically isolated solutions is finite.
\end{proof}

\paragraph{Remark.}
All the models considered in this paper have moments $\mu$ which correspond to
a polynomial map.  However, for some models (e.g., exponential families),
$\mu$ will not be a polynomial map, but rather, a general analytic function.
In this case, \refthm{checker} holds with one modification to (ii).
If $\CheckIdentifiability$ returns ``yes'', then we have the
following weaker guarantee in place of (ii):
$\eqclass{\Thetamax}(\thetatrue)$ is \emph{countable} (but not necessarily
finite) for all $\thetatrue \in \Thetamax$.
The above proof does not require the fact that $\mu$ is a polynomial map
except in the invocation of B\'ezout's Theorem.
In place of B\'ezout's Theorem, we use the following argument.
If $\eqclass{\Thetamax}(\thetatrue)$ is uncountable, then it contains a
limit point $\theta^* \in \eqclass{\Thetamax}(\thetatrue)$; thus for any
small enough neighborhood $N(\theta^*)$ of $\theta^*$, there is some
$\theta \in \eqclass{\Thetamax}(\thetatrue) \cap N(\theta^*)$.
This contradicts (i) as applied to $\theta^*$, and thus we
conclude that $\eqclass{\Thetamax}(\thetatrue)$ is countable.

\Section{more}{Additional results from the identifiability checker}

\paragraph{PCFG models with $d < k$.}

The PCFG models that we've considered so far assume that the number of words
$d$ is at least the number of hidden states $k$, which is a realistic
assumption for natural language.  However, there are applications, e.g.,
computational biology, where the vocabulary size $d$ is relatively small.
In this regime, identifiability becomes trickier because the data doesn't
reveal as much about the hidden states, and brings us closer to the boundary
between identifiability and non-identifiability.  In this section, we consider the $d <
k$ regime.

The following table gives additional identifiability results from
$\CheckIdentifiability$ for values of $d$, $k$, and $L$ where $d < k$
(recall that the results reported in \refsec{results} only considered values where $d \geq k$).
In each cell, we show the $(k,d,L)$ values for which
$\CheckIdentifiability$ returned ``yes''; the values checked were
$k \in \{3,4,\dotsc,8\}$, $d \in \{2,\dotsc,k-1\}$, $L \in
\{3,4,\dotsc,9\}$.

\begin{center}
\begin{tabular}{c|c|c|c|c|c|c|}
\cline{2-7}
&
$\phi_\Pairs$ &
$\phi_\AllPairs$ &
$\phi_\ThinTriples{e_1}$ &
$\phi_\Triples$ &
$\phi_\AllThinTriples{e_1}$ &
$\phi_\AllTriples$ \\
\hline
\multicolumn{1}{|c|}{PCFG} &
\multicolumn{6}{|c|}{None} \\
\hline
\multicolumn{1}{|c|}{PCFG-I} &
None &
\tcell{
$(3,2,\geq6)$ \\
$(4,2,\geq8)$ \\
$(4,3,\geq5)$ \\
$(5,3,\geq6)$ \\
$(5,4,\geq4)$ \\
$(6,3,\geq7)$ \\
$(6,4,\geq5)$ \\
$(6,5,\geq4)$ \\
$(7,3,\geq8)$ \\
$(7,4,\geq6)$ \\
$(7,5,\geq5)$ \\
$(7,6,\geq4)$
} &
None &
\tcell{
$(5,4,\geq4)$ \\
$(6,5,\geq4)$ \\
$(7,5,\geq4)$ \\
$(7,6,\geq4)$
} &
\multicolumn{2}{|c|}{\tcell{
$(3,2,\geq5)$ \\
$(4,2,\geq6)$ \\
$(4,3,\geq4)$ \\
$(5,2,\geq7)$ \\
$(5,\geq3,\geq4)$ \\
$(6,2,\geq8)$ \\
$(6,3,\geq5)$ \\
$(6,\geq4,\geq4)$ \\
$(7,2,\geq9)$ \\
$(7,3,\geq5)$ \\
$(7,\geq4,\geq4)$
}} \\
\hline
\multicolumn{1}{|c|}{PCFG-IE} &
None &
\tcell{
$(3,2,\geq6)$ \\
$(4,2,\geq8)$ \\
$(4,3,\geq5)$ \\
$(5,3,\geq6)$ \\
$(5,4,\geq5)$ \\
$(6,3,\geq7)$ \\
$(6,4,\geq5)$ \\
$(6,5,\geq4)$ \\
$(7,3,\geq8)$ \\
$(7,4,\geq6)$ \\
$(7,5,\geq5)$ \\
$(7,6,\geq4)$
} &
\tcell{
$(5,4,\geq4)$ \\
$(6,5,\geq4)$ \\
$(7,5,\geq5)$ \\
$(7,6,\geq4)$
} &
\tcell{
$(4,3,\geq4)$ \\
$(5,4,\geq4)$ \\
$(6,\geq4,\geq4)$ \\
$(7,\geq5,\geq4)$ \\
} &
\tcell{
$(3,2,\geq5)$ \\
$(4,2,\geq6)$ \\
$(4,3,\geq4)$ \\
$(5,2,\geq7)$ \\
$(5,3,\geq5)$ \\
$(5,4,\geq4)$ \\
$(6,2,\geq8)$ \\
$(6,3,\geq5)$ \\
$(6,\geq4,\geq4)$ \\
$(7,2,\geq9)$ \\
$(7,3,\geq5)$ \\
$(7,\geq4,\geq4)$
} &
\tcell{
$(3,2,\geq5)$ \\
$(4,2,\geq6)$ \\
$(4,3,\geq4)$ \\
$(5,2,\geq7)$ \\
$(5,\geq3,\geq4)$ \\
$(6,2,\geq8)$ \\
$(6,3,\geq5)$ \\
$(6,\geq4,\geq4)$ \\
$(7,2,\geq9)$ \\
$(7,3,\geq5)$ \\
$(7,\geq4,\geq4)$
} \\
\hline
\end{tabular}
\end{center}

\paragraph{Fixed topology models.}

We now present some results for latent class models (LCMs) and hidden Markov
models (HMMs).  While identifiability for these models are more developed than
for parsing models, we show that the identifiability checker can refine the
results even for the classic models.

The parameters of an HMM are $\theta = (\pi, T, O)$, where $\pi \in \R^k$
specifies the initial state distribution, $T \in \R^{k \times k}$ specifies
the state transition probabilities, and $O \in \R^{d \times k}$ specifies
the emission distributions.
The probability over a sentence $\bx$ is:
\begin{equation} \label{eq:hmm}
\BP_\theta(\bx)
= \bone^\top
T\diag(O^\top x_L)
\dotsb
T\diag(O^\top x_2)
T\diag(O^\top x_1)
\pi
.
\end{equation}
The parameters of an LCM are $\theta = (\pi, O)$---the same as that of an
HMM except with $T \equiv I$.
The probability over a sentence $\bx$ is also given by \eqref{eq:hmm} (with
$T = I$).

The following table summarizes some identifiability results obtained by
$\CheckIdentifiability$ (for $d \geq k$); these results have all been
proven analytically in previous work (e.g.,
\cite{kruskal77three,chang96ident,mossel06hmm,hsu09spectral,allman09identifiability})
except for the identifiability of HMMs from $\phi_\AllPairs$.

\begin{center}
\begin{tabular}{c|c|c|c|c|c|c|}
\cline{2-7}
&
$\phi_\Pairs$ &
$\phi_\AllPairs$ &
$\phi_\ThinTriples{e_1}$ &
$\phi_\Triples$ &
$\phi_\AllThinTriples{e_1}$ &
$\phi_\AllTriples$ \\
\hline
\multicolumn{1}{|c|}{LCM} &
\multicolumn{2}{|c|}{No} &
\multicolumn{4}{|c|}{\tcell{Yes iff $L\geq3$}} \\
\hline
\multicolumn{1}{|c|}{HMM} &
No &
\multicolumn{5}{|c|}{\tcell{Yes iff $L\geq3$}} \\
\hline
\end{tabular}
\end{center}

It is known that LCMs are not identifiable from $\phi_\AllPairs$ for any
value of $L$~\cite{chang96ident}.
However, LCMs constitute a subfamily of HMMs arising from a measure zero
subset of the HMM parameter space.
Therefore the identifiability of HMMs from $\phi_\AllPairs$ (for $L\geq3$)
does not contradict this result.
The result does not appear to be covered by application of Kruskal's
theorem in previous work \cite{allman09identifiability}, so we prove the
result rigorously below.

It can be checked using \eqref{eq:hmm} that
\begin{align*}
\E_\theta[ \phi_{12}(\bx) ] & = O \diag(\pi) T^\top O^\top \\
\E_\theta[ \phi_{34}(\bx) ] & = O \diag(T\pi) T^\top O^\top
.
\end{align*}
Let $M_1 \eqdef O$, $M_2 \eqdef O T \diag(\pi)$, and $D \eqdef \diag(T\pi)
\diag(\pi)^{-1}$.
Provided that
\begin{enumerate}
\item $\pi > 0$,
\item $O$ has full column rank,
\item $T$ is invertible,
\item the ratios of probabilities $(T\pi)_i / \pi_i$, ranging over $i \in [k]$, are distinct
\end{enumerate}
(all of which are true for all but a measure zero set of parameters in
$\Theta$), the matrices $M_1$ and $M_2$ have full column rank and the
diagonal matrix $D$ has distinct diagonal entries.
Therefore \reflem{decompose} can be applied with $X = \E_\theta[
\phi_{12}(\bx) ] = M_1M_2^\top$ and $Y = \E_\theta[ \phi_{34}(\bx) ] =
M_1DM_2^\top$ to recover $M_1 = O$.
It is easy to see that $\pi$ and $T$ can also easily be recovered.

Note that the fourth condition above, that $T\pi$ be entry-wise distinct
from $\pi$, is violated when a LCM distribution is cast as an HMM
distribution (by setting $T = I$ so $T\pi = \pi$).
However, the set of HMM parameters satisfying this equation is a measure
zero set.

\paragraph{Discussion.}

$\CheckIdentifiability$ tests for local identifiability.
If it finds that a model family is not locally identifiable, then it is not
globally identifiable.
However the inverse claim is not necessarily true: if it finds that a model family is
locally identifiable, it is not necessarily globally identifiable.
\refthm{checker} provides the somewhat weaker guarantee that a restricted
model family is globally identifiable, where the equivalence classes
$\eqclass{\Theta\setminus\sE}(\thetatrue)$ are only taken with respect to
a subset $\Theta\setminus\sE \subseteq \Theta$ of the parameter space.
However, there is a gap between this property (which is with respect to $\Theta\setminus\sE$)
and true global identifiability (which is with respect to $\Theta$).

On the other hand, having explicit estimators guarantees us
proper global identifiability with respect to the original model family $\Theta$.
In fact, the exceptional set $\sE$ can typically be characterized explicitly.
For instance, in the case of PCFG-IE, the set $\Theta\setminus\sE$ contains
those $\theta = (\pi,T,O)$ that satisfy full rank conditions:
\begin{align}
\Theta\setminus\sE &= \{ (\pi,T,O) : \pi \succ 0, \text{$T$ is invertible}, \text{$O$ has full column rank} \}.
\end{align}

Additionally, the explicit estimators also provides an explicit
characterization of the elements in the equivalence class
$\sS_\Theta(\theta_0)$ for each $\theta_0 \in \Theta\setminus\sE$:
the set $\sS_\Theta(\theta_0)$ contains exactly $k!$ elements
corresponding to permutation of the hidden states.
Specifically,
\begin{align}
\sS_\Theta((\pi,T,O)) &= \{ (\Pi^{-1} \pi, \Pi^{-1} T \Pi, O \Pi) : \text{$\Pi$ is a permutation matrix}.
\end{align}
Note that this is shaper than \refthm{checker}, which only says that the
equivalence classes have to be finite.

\Section{dynamicPrograms}{Dynamic programs}

For a sentence of length $L$, the number of parse trees is exponential in $L$.
Therefore, dynamic programming is
often employed to efficiently compute expectations over the parse trees,
the core computation in the E-step of the EM algorithm.
In the case of PCFG, this dynamic program is referred to as the CKY
algorithm, which runs in $O(L^3 k^3)$ time, where $k$ is the number of hidden states.
For simple dependency models, a $O(L^3)$ dynamic program was developed by
\cite{eisner00cubic}.
At a high-level, the states of the dynamic program in both cases are
the spans $\Span{i}{j}$ of the sentence (and for the PCFG, the these states
include the hidden states $z_{\Span{i}{j}}$ of the nodes).

In this paper, we need to compute (i) the Jacobian matrix for checking
identifiability (\refsec{identifiabilityChecker}) and (ii) the mixing matrix for
recovering compound parameters (\refsec{unmixGeneral}).  Both computations can
be performed efficiently with a modified version of the classic dynamic programs,
which we will describe in this section.

\Subsection{dynamicProgramJacobian}{Computing the Jacobian matrix}

Recall that the $j$-th row of the Jacobian matrix $J$ is (the transpose of) the
gradient of $h_j(\theta) = \mu_j(\theta) - \mu_j(\thetatrue)$.
Specifically, entry $J_{ji}$ is 
the derivative of the $j$-th moment with respect to the $i$-th parameter:
\begin{align}
J_{ji}
&= \frac{\partial h_j(\theta)}{\partial \theta_i} \\
&= \frac{\partial {\E}_{\theta}[\phi_j(\bx)]}{\partial \theta_i} \\
&= \sum_{\bx, z} \frac{\partial p_\theta(\bx, z)}{\partial \theta_i} \phi_j(\bx).
\end{align}

We can encode the sum over the exponential set of possible sentences $\bx$ and
parse trees $z$ using a directed acyclic hypergraph so that each hyperpath through
the hypergraph corresponds to a $(\bx,z)$ pair.
Specifically, a {\em hypergraph} consists of the following:
\begin{itemize}
\item a set of nodes $\sV$ with a designated start node $\StartNode \in \sV$
and an end node $\EndNode \in \sV$, and
\item a set of hyperedges $\sE$ where each hyperedge $e \in \sE$
has a source node $e.a \in \sV$ and a pair of target nodes
$(e.b, e.c) \in \sV \times \sV$ (we say that $e$ connects $e.a$ to $e.b$ and $e.c$)
and an index $e.i \in [n]$ corresponding to a component of
the parameter vector $\theta \in \R^n$.
\end{itemize}

Define a {\em hyperpath} $P$ to be a subset of the edges $\sE$ such that:
\begin{itemize}
\item $(\StartNode, a, b) \in P$ for some $a,b \in \sV$;
\item if $(a,b,c) \in P$ and $b \neq \EndNode$, then $(b, d, e) \in P$ for some $d,e \in \sV$; and
\item if $(a,b,c) \in P$ and $c \neq \EndNode$, then $(c, d, e) \in P$ for some $d,e \in \sV$.
\end{itemize}

Each hyperpath $P$, encoding $(\bx, z)$, is associated with a probability
equal to the product of all of the parameters on that hyperpath:
\begin{align}
p_\theta(\bx, z) = p_\theta(P) = \prod_{e \in P} \theta_{e.i}.
\end{align}
In this way, the hypergraph compactly defines a distribution over exponentially
many hyperpaths.

Now, we assume that each moment $\phi_j(\bx)$ corresponds to a function $f_j :
\sE \mapsto \R$ mapping each hyperedge $e$ to a real number so that
the moment is equal to the product over function values:
\begin{align}
\phi_j(\bx) &= \prod_{e \in P} f_j(e),
\end{align}
where $P$ is any hyperpath that encodes the sentence $\bx$ and some parse tree $z$
(we assume that the product is the same no matter what $z$ is).

Now, let us write out the Jacobian matrix entries in terms of hyperpaths:
\begin{align}
J_{ji}
& = \sum_P \sum_{e_0 \in P} \frac{\partial \theta_{e_0.i}}{\partial \theta_i} \prod_{e \in P, e \neq e_0} \theta_{e.i} f_j(e).
\end{align}

The sum over hyperpaths $P$ can be computed efficiently as follows.
For each hypergraph node $a$,
we compute an inside score $\alpha(a)$,
which sums over all possible partial hyperpaths terminating at the target node,
and an outside score $\beta(a)$,
which sums over all possible partial hyperpaths from the source node:
\begin{align}
\alpha(a) &\eqdef \sum_{e \in \sE : e.a = a} \theta_{e.i} \alpha(e.b) \alpha(e.c), \\
\beta(a) &\eqdef
\sum_{e \in \sE : e.b = a} \theta_{e.i} \alpha(e.c) \beta(e.a)
\sum_{e \in \sE : e.c = a} \theta_{e.i} \alpha(e.b) \beta(e.a).
\end{align}

The Jacobian entry $J_{ji}$ can be computed as follows:
\begin{align}
J_{ji} &= \sum_{e \in \sE} \beta(e.a) \alpha(e.b) \alpha(e.c) \1[i = e.i].
\end{align}

\paragraph{Example: PCFG.}

For a PCFG, nodes $\sV$ have the form $(i,j,s) \in [L] \times [L] \times [k]$,
corresponding to a hidden state $s$ over span $\Span{i}{j}$.
For each hidden state $s$, we have a hyperedge $e$ connecting $e.a =
\StartNode$ to $e.b = (s, 0, L)$
and $e.c = \EndNode$; this hyperedge has parameter index $e.i$ corresponding to $\pi_s$.
For each span $\Span{i}{j}$ with $j-i > 1$,
split point $i < m < j$, and hidden states $s_1,s_2,s_3 \in [k]$,
$\sE$ contains a hyperedge $e$ connecting
$e.a = (i, j, s_1)$ to $e.b = (i, m, s_2)$ and $e.c = (m, j, s_3)$;
the parameter index $e.i$ corresponds to the binary production 
$B_{(s_2 \otimes_k s_3) s_1}$.
For each span $\Span{i-1}{i}$, hidden state $s \in [k]$ and word $x \in [d]$,
we have a hyperedge $e$ connecting $e.a = (i-1,i,s)$ to $e.b = \EndNode$ and $e.c = \EndNode$
with parameter index $e.i$ corresponding to the emission $O_{x s}$.

The moments can be encoded as follows:
For example, if $\phi_j(\bx) = \1[x_i = t]$,
then we define $f_j(e)$ to be 0 if the source node corresponds to position $i$
($e.a = (i-1,i,s)$) and the parameter index $e.i$ does not correspond
to $O_{t s}$ for some $s \in [k]$, and 1 otherwise.
In this way, $\prod_{e \in P} f_j(e)$ is zero if $P$ encodes
a sentence with $x_i \neq t$.

Higher-order moments simply correspond to hyperedge-wise multiplication of
these first-order moments.  For example,
if $\phi_{j_1}(\bx) = \1[x_{i_1} = t_1]$
and $\phi_{j_2}(\bx) = \1[x_{i_2} = t_2]$,
then the second-order moment
$\phi_{j}(\bx) = \1[x_{i_1} = t_1, x_{i_2} = t_2]$
corresponds to $f_j(e) = f_{j_1}(e) f_{j_2}(e)$.

\Subsection{dynamicProgramMixing}{Computing the mixing matrix}

Recall that the mixing matrix $M$ includes a row for each observation matrix $o \in
\sO$ and a column for each compound parameter $p \in \sP$.
Assuming a uniform distribution over topologies,
computing each entry of $M$ reduces to counting the number of topologies $t$
consistent with a particular compound parameter $\Psi_p$:
\begin{align}
M_{op}
&= \bP(\Psi_{o,\Tree} = \Psi_p) \\
&= |\Trees|^{-1} \sum_t \1[\Psi_{o,t} = \Psi_p].
\end{align}

\FigTop{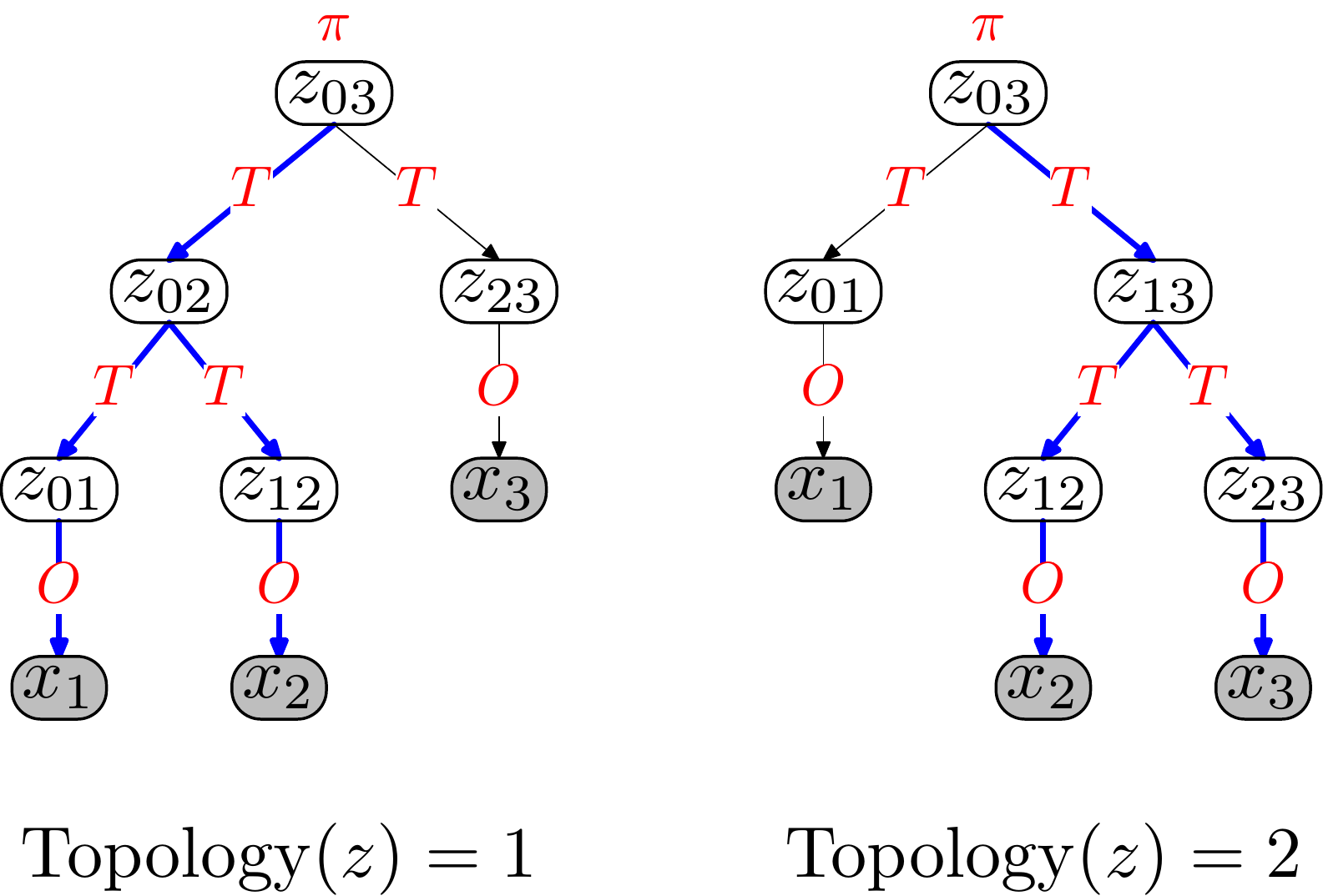}{0.4}{backboneExample}{An example of a backbone structure
in blue corresponding to the compound parameter $O T \diag(T \pi) T^\top O^\top$,
which appears in two different topologies, for two observation matrices,
$\phi_{12}$ and $\phi_{23}$, respectively.}

First, we will characterize the set of compound parameters graphically in terms
of backbone structures.
As an example, consider the PCFG-IE model and the observation matrix $\phi_{12}$
($o = 12$) corresponding to the marginal distribution over the first two words
of the sentence.  Given a topology $t$, consider starting at the root,
descending to the lowest common ancestor of $x_1$ and $x_2$, and then following
both paths down to $x_1$ and $x_2$, respectively.  We refer to this traversal
as the {\em backbone structure} with respect to topology $t$ and observation
matrix $\phi_{12}$.  See \reffig{backboneExample} for an example of the
backbone structure, outlined in blue.

Note that the compound parameter
$\Psi_{12,t}(\theta) = \E_{\theta}[\phi_{12}(\bx) \mid \Tree = t]$
can be written as a product over the parameter matrices, one for each edge of
the backbone structure.
For \reffig{backboneExample}, this would yield
\begin{align}
\Psi_{12,1}(\theta) = O T \diag(T \pi) T^\top O^\top.
\end{align}
For general trees, we would have
\begin{align}
\label{eqn:pairsPath}
\Psi_{12,t}(\theta) = O T^{n_1} \diag(T^{n_3} \pi) (T^\top)^{n_2} O^\top.
\end{align}
for some positive integers $n_1,n_2,n_3$ corresponding to the number of edges (in $t$)
from the common node to the preterminal node $z_{01}$, the preterminal node $z_{12}$,
and the root $z_{0L}$, respectively.

Note that the compound parameter does not depend on the structure of $t$ outside
the backbone---that part of the topology is effectively marginalized out---so
the compound parameter $\Psi_{12,t}(\theta)$ will be identical for all
topologies sharing that same backbone structure.  Therefore, there are only a
polynomial number of compound parameters despite an exponential number of
topologies $t$.\footnote{One might also see why the unmixing technique does not directly
apply to the PCFG-I model, where $T$ is replaced with $T_1$ for left edges and
$T_2$ for right edges. In that case, there are many backbone structures
(and thus more compound parameters)
due to the different interleavings of left and right edges.}

We define a dynamic program that recursively computes $M_{op}$ for the PCFG-IE
model under a fixed second-order observation matrix $\phi_{i_0 j_0}$.
Specifically, for each span $\Span{i}{j}$
define $H(i,j)$ to be the set of pairs $\inner{t}{n}$
where $t$ is a partial backbone structure $t$ and $n$ is the number of
partial topologies over span $\Span{i}{j}$ 
which are consistent with $t$.

In the base case $H(i-1,i)$, if $i$ is either of the designated leaf positions
defined by the observation matrix
($i_0$ or $j_0$), then we return the single-node backbone structure $\bullet$;
otherwise, we return the null backbone structure $\nil$:
\begin{align}
H(i-1, i) &= \begin{cases} \{ \inner{\nil}{1} \} & \text{if $i = i_0$ or $i = j_0$} \\ \{ \inner{\bullet}{1} \} & \text{otherwise}. \end{cases}
\end{align}

In the recursive case $H(i,j)$, we consider all split points $m$, partial backbones
$t_1$ and $t_2$ from $H(i,m)$ and $H(m,j)$, respectively, and create a new tree
with $t_1$ and/or $t_2$ as the subtrees if they are not null:
\begin{align}
H(i,j) &=
  \bigcup^+_{i < m < j}
  \bigcup^+_{\inner{t_1}{n_1} \in H(i, m)}
  \bigcup^+_{\inner{t_2}{n_2} \in H(m, j)}
    \inner{\Combine(t_1, t_2)}{n_1 n_2}, \\
\Combine(t_1, t_2) &=
\begin{cases} 
(T:t_1, T:t_2) & \text{if $t_1 \neq \nil$ and $t_2 \neq \nil$}, \\
T:t_1 & \text{if $t_1 \neq \nil$}, \\
T:t_2 & \text{if $t_2 \neq \nil$}, \\
\nil & \text{otherwise}.
\end{cases}
\end{align}
Here, we use the notation $\bigcup^+$ to denote a multi-set union:
$\{ \inner{t}{n_1} \} \cup^+ \{ \inner{t}{n_2} \} = \{ \inner{t}{n_1+n_2} \}$.
In this notation, the backbone structure in \reffig{backboneExample}
would be represented as $T:(T:O:\bullet,T:O:\bullet)$, which can be easily converted
to the compound parameter $O T \diag(T \pi) T^\top O^\top$.

For third-order observation matrices (e.g., $\phi_{i_0 j_0 k_0 \eta}$),
we add an additional case to $H(i-1,i)$ to return $\inner{\circ}{1}$ if $i = k_0$;
note that $k_0$ is represented by a special node $\circ$ because that observation
is projected using $\eta$.
The first case of $\Combine(t_1,t_2)$ undergoes one change:
if $t_2$ is a chain ending in $\circ$, then we return $(T:t_2, T:t_1)$.
The reason for this is best demonstrated by an example:
consider topology 1 in \reffig{backboneExample},
and the two observation matrices $\phi_{132\eta}$ and $\phi_{231\eta}$.
Without the reordering, we would have the backbone structure:
$(T:(T:\bullet,T:\circ),T:\bullet)$ and
$(T:(T:\circ,T:\bullet),T:\bullet)$.
However, they have the same compound parameter
$O T \diag(T^\top O^\top \eta) T^\top \diag(\pi) T O$.
This is because the contribution of a subtree ending in $\circ$
is simply a diagonal matrix ($\diag(T^\top O^\top \eta)$ in this case)
which is applied on the hidden state regardless of whether it came from the
left or right side.

\end{document}